%% file: main.tex
\documentclass{ecai} 

\usepackage{latexsym}
\usepackage{amssymb}
\usepackage{amsmath}
\usepackage{amsthm}
\usepackage{booktabs}
\usepackage{enumitem}
\usepackage{graphicx}
\usepackage{color}
\usepackage[caption=false]{subfig}
\usepackage{nicefrac}
\usepackage{multirow}
\usepackage{fancyhdr}  

\newcommand{\BibTeX}{B\kern-.05em{\sc i\kern-.025em b}\kern-.08em\TeX}

\newcommand{\approach}[0]{\textsc{ST-SampleNet}}
\newcommand{\resnet}[0]{\textsc{ResNet}}
\newcommand{\transformer}[0]{\textsc{Encoder}}

\begin{document}


\begin{frontmatter}

\paperid{2055}

\title{Spatially Constrained Transformer with Efficient Global Relation Modelling for Spatio-Temporal Prediction}

\author[A]{\fnms{Ashutosh}~\snm{Sao}\thanks{Corresponding Author. Email: sao@L3S.de}}
\author[A]{\fnms{Simon}~\snm{Gottschalk}}

\address[A]{L3S Research Center, Leibniz University Hannover}

\input{01_abstract}

\end{frontmatter}

\thispagestyle{fancy}
\fancyhf{}  
\fancyfoot[C]{\textcopyright\ Ashutosh Sao and Simon Gottschalk, 2024. The definitive, peer reviewed and edited version of this Article is published in 27th European Conference on Artificial Intelligence, 19–24 October 2024, Santiago de Compostela, Spain – Including 13th Conference on Prestigious Applications of Intelligent Systems (PAIS 2024), Ulle Endriss, Francisco S. Melo, Kerstin Bach, Alberto Bugarín-Diz, José M. Alonso-Moral, Senén Barro, Fredrik Heintz, ISBN: 978-1-64368-548-9 (online), 2781–2789, 2024, and DOI: 10.3233/FAIA240813. 
}

\input{02_introduction}

\input{03_problem}

\input{04_approach}

\input{05_setup}

\input{06_result}

\input{07_relatedwork}

\input{08_conclusion}

\begin{ack}
This work was partially funded by the Federal Ministry for Economic Affairs and Climate Action (BMWK), Germany (“ATTENTION!”, 01MJ22012D), and by the Federal Ministry for Digital and Transport (BMDV), Germany (“MoToRes”, 19F2271A).
\end{ack}

\bibliography{mybibfile}

\end{document}

%% file: 01_abstract.tex
\begin{abstract}
Accurate spatio-temporal prediction is crucial for the sustainable development of smart cities. However, current approaches often struggle to capture important spatio-temporal relationships, particularly overlooking global relations among distant city regions.
Most existing techniques predominantly rely on Convolutional Neural Networks (CNNs) to capture global relations. However, CNNs exhibit neighbourhood bias, making them insufficient for capturing distant relations.
To address this limitation, we propose \approach{}, a novel transformer-based architecture that combines CNNs with self-attention mechanisms to capture both local and global relations effectively. Moreover, as the number of regions increases, the quadratic complexity of self-attention becomes a challenge. To tackle this issue, we introduce a lightweight region sampling strategy that prunes non-essential regions and enhances the efficiency of our approach.
Furthermore, we introduce a spatially constrained position embedding that incorporates spatial neighbourhood information into the self-attention mechanism, aiding in semantic interpretation and improving the performance of \approach{}.
Our experimental evaluation on three real-world datasets demonstrates the effectiveness of \approach{}. Additionally, our efficient variant achieves a $40\%$ reduction in computational costs with only a marginal compromise in performance, approximately $1\%$.
\end{abstract}

%% file: 02_introduction.tex
\section{Introduction}
\label{sec:intro}

Spatio-temporal prediction aims to predict future spatio-temporal dynamics based on historical data. This prediction task is commonly categorised into two domains: road network-level predictions \cite{fogs,lsgcn,BiRNet}, often modelled as graph-based problems, and region-level predictions \cite{st_gsp,deep_crowd}, typically approached as grid-based problems. 
This paper addresses region-level prediction, emphasising its significance for city planners, administrators, and ride-hailing companies. Focusing on this level is crucial for informed decision-making in smart city planning and development \cite{smartcity_plan}. 
Accurate region-level prediction enables one to view city-wide traffic dynamics, which helps in the sustainable development of cities to provide better access to transport and public services. 

For accurate spatio-temporal prediction, three essential spatio-temporal relations need to be addressed: 
\begin{itemize}
    \item \textit{Local Spatial Relations} between closed neighbouring regions. For example, congestion in one region often affects adjacent regions due to the interconnectedness of road networks.

    \item \textit{Global Spatial Relations} between distant regions. For instance, distant residential areas often share similar traffic patterns, such as peak traffic during morning and evening rush hours.

    \item \textit{Temporal Relations} between input time intervals. For example, traffic decreases in office areas in late hours as people return home.
\end{itemize}

Existing approaches predominantly focus on capturing local spatial relations by utilising CNN-based networks \cite{st-resnet,deepstn+,st3dmddn} and temporal relations through LSTMs/GRUs \cite{stdn,deep_crowd} or Transformers \cite{st_gsp,bdstn}, and often overlook global relations by solely relying on CNNs to capture them. 
CNNs inherently exhibit a neighbour bias, meaning they are better at capturing local patterns and relationships within a spatial context. However,
their ability to capture global relations is limited due to their localised receptive fields and parameter sharing, which prioritise nearby information over distant contexts.
To overcome these limitations, we propose \approach{}, which, besides capturing local and temporal relations effectively, also captures global relations by employing the self-attention mechanism \cite{attention}. 

However, as the number of regions increases either by expanding the observation area (larger city or state) or increasing the granularity (smaller region size), the quadratic complexity of self-attention becomes the major bottleneck for training and inference.

\begin{figure}[t]
    \centering
    \subfloat[Hannover Traffic $8-9$ am.\label{fig:traffic}]{%
  \includegraphics[width=0.45\linewidth]{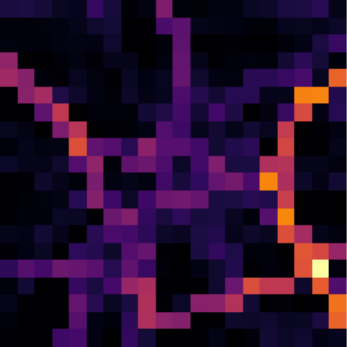}%
}%
\hfill
\subfloat[Corresponding Attention Map.\label{fig:attention}]{%
  \includegraphics[width=0.45\linewidth]{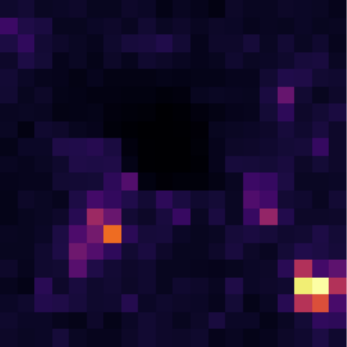}%
}
\caption{Hannover traffic between $8-9$ am of a working day and the corresponding attention map from \approach{}. The sparse attention demonstrates the dependency on only a few regions for prediction.}
\label{fig:motivation}
\end{figure}


We observe that different regions hold varying importance at different time intervals, making it unnecessary to include all regions in each prediction. For example, if we know the traffic conditions on a particular highway, we can infer the traffic conditions for other regions on the same highway, thus eliminating the need to include them all. Figure \ref{fig:traffic} illustrates the traffic in the German city Hannover between $8-9$ am on a weekday. High traffic volumes towards the east, north-west, and south correspond to highways, while moderate to low traffic in central areas correspond to residential, commercial, and recreational areas. Figure \ref{fig:attention} shows the corresponding attention map generated by \approach{}. 
It can be seen that \approach{} requires only a limited amount of regions of varying traffic levels (from low to high) to predict the city-wide traffic conditions indicating that other regions are redundant, i.e., their traffic can be inferred using the information from the important regions.

Therefore, we propose a lightweight region sampling strategy that prunes non-essential regions at different intervals and addresses the quadratic complexity inherent in self-attention mechanisms.
For sampling, we utilised Gumbel-Softmax \citep{gumbel_softmax}, which has a two-fold advantage. Firstly, it is differentiable, thus making end-to-end training possible. Secondly, it introduces stochasticity to the sampling process, occasionally selecting non-important regions, thereby enhancing the model's generalisation capability through exploratory behaviour.

Additionally, we introduce a spatially constrained and learnable position embedding (SCPE) that integrates neighbourhood information in a hierarchical manner, such as locality, city, and state levels and enhances semantic interpretability. This approach constrains the self-attention mechanism to prioritise immediate neighbours, while simultaneously the learnable structure of the embedding also provides the flexibility to capture relations among distant neighbours. Consequently, SCPE not only comprehends spatial dependencies across different distances but also offers semantic interpretability.

Thus, overall, our contributions are as follows:
\begin{itemize}
    \item We propose \approach{}\footnote{https://github.com/ashusao/ST-SampleNet}, a novel spatio-temporal prediction model that effectively captures crucial spatio-temporal relations, including global relations, often neglected in prior approaches.

    \item We propose a lightweight region sampling strategy that prunes the non-important regions and makes \approach{} efficient.

    \item We propose a learnable and interpretable position embedding that embeds spatial neighbourhood information.
    
    \item Experiments on three real-world datasets prove the efficiency and effectiveness of \approach{}.
\end{itemize}

%% file: 03_problem.tex
\section{Preliminaries}

In this section, we formally define the problem of spatio-temporal prediction and the external features used in \approach{}.

\subsection{Problem Statement}

\paragraph{Region} Following \cite{deepst}, we divide a city into $H \times W$ uniform grids, each representing a distinct region denoted as $r^n$, $n \in [1, H \times W]$.

\paragraph{Spatio-Temporal Image} During each time interval $t$, we collect spatio-temporal measurements \cite{deepst} (e.g., inflow/outflow) for individual regions. The measurements for each region $r^n$ are combined to create a single-channel image, and multiple such images are stacked together to construct a spatio-temporal image (referred to as an image) \cite{periodic_crn},  denoted as $\mathcal{X}_t \in \mathbb{R}^{M \times H \times W}$. Here, $M$ signifies the total number of distinct measurements.

\paragraph{Spatio-Temporal Prediction} The objective is to forecast the image at the subsequent time interval $\mathcal{X}_{t+1}$ given a sequence of historical images $\langle \mathcal{X}_t, \mathcal{X}_{t-1}, \dots \rangle$.

\subsection{External Features}
 Spatio-temporal measurements depend on the region's semantic attributes (e.g., residential or commercial nature) and temporal factors. For instance, peak traffic in residential and office areas can be seen during morning and evening rush hours. Hence, we incorporate these two factors as external features in our model.

\paragraph{Semantic Features} To capture the semantic aspects of a region, we utilise Points of Interest (POIs). For each region $r^n$, the count of POIs in specific categories $p$ (e.g. count of residential buildings) is quantified as follows:
\begin{align}
\label{eq:poi_count}
x^{p, r^n} = | \{ q\ |\ q \in \mathcal{Q}^{p}\ \textnormal{s.t.}\ g_q \in r^n \}|,
\end{align}
where $g_{q} \in r^n$ indicates that the POI with location $g_q$ lies within region $r^n$ and $\mathcal{Q}^p$ is the set of all POIs of type $p$ in the city. Aggregated POI counts form single-channel images for each category, stacked to create the multi-dimensional POI image $\mathcal{X}^{poi} \in \mathbb{R}^{P \times H \times W}$, with $P$ denoting the total number of distinct categories.

\paragraph{Temporal Features} Regarding temporal information, we employ one-hot encoding to capture day-of-the-week information $x^{dow}_t$ and weekend status  $x^{we}_t$. Due to the cyclic nature of the time of day ($tod$), sinusoidal encoding is applied:
\begin{equation}
x^{tod}_t = [ \sin \left( \frac{2 \pi t'}{1440} \right), \cos \left( \frac{2 \pi t'}{1440} \right)],
\end{equation}
where $t'$ represents the time elapsed in minutes from midnight until the start of time interval $t$. The three-time encodings are concatenated to form the feature vector $\mathcal{X}^{time}_t = [x^{dow}_t; x^{we}_t; x^{tod}_t] \in \mathbb{R}^{10}$, encompassing comprehensive temporal features in our model.

%% file: 04_approach.tex
\section{Approach}
In this section, we present the \approach{} approach, covering its architecture, spatially constrained position embedding, region sampling strategy, and training overview.

\subsection{Architecture}
\begin{figure*}
    \centering
    \includegraphics[width=1.0\textwidth]{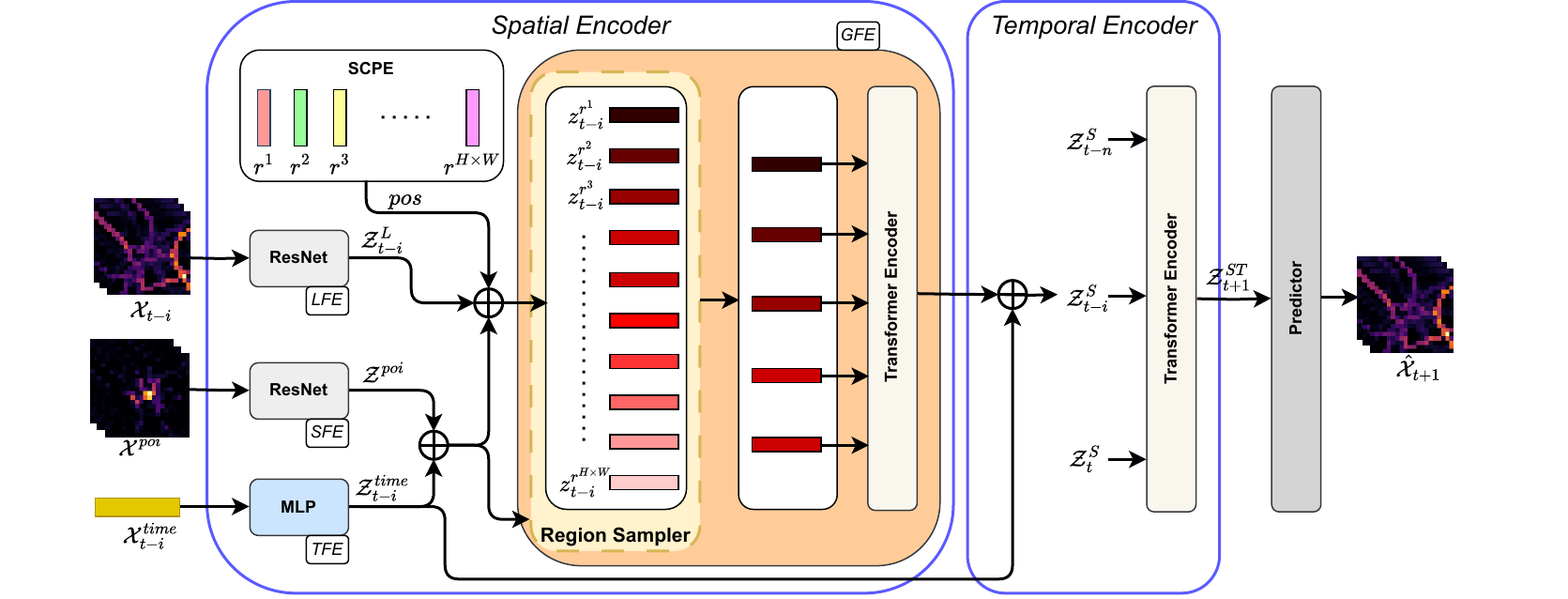}
    \caption{\approach{} architecture has three main components: (i) \textit{Spatial Encoder} -- learns the spatial dependency amongst different regions, (ii) \textit{Temporal Encoder} -- learns the temporal dependency amongst input time intervals and (iii) \textit{Predictor} -- makes the final predictions.}
    \label{fig:arch}
\end{figure*}

Figure \ref{fig:arch} illustrates the architecture of \approach{}. It consists of three primary components: the \textit{Spatial Encoder}, the \textit{Temporal Encoder}, and the \textit{Predictor}.
For a given timestamp $t-i$, the \textit{Spatial Encoder} takes three inputs: the spatio-temporal image $\mathcal{X}_{t-i}$, the semantic features of the city $\mathcal{X}^{poi}$, and the temporal features $\mathcal{X}^{time}_{t-i}$. It processes these inputs to generate the spatial representation $\mathcal{Z}^{S}_{t-i}$.

Subsequently, the \textit{Temporal Encoder} operates on the spatial representations of input intervals $\langle \mathcal{Z}^{S}_t, \mathcal{Z}^{S}_{t-1}, \dots, \mathcal{Z}^{S}_{t-n} \rangle$, producing the final representation $\mathcal{Z}^{ST}_{t+1}$ of the subsequent time interval $t+1$. 
This final representation is then utilised by the \textit{Predictor} to make predictions ($\hat{\mathcal{X}}_{t+1}$). Further details regarding each component are provided below.

\subsubsection{Spatial Encoder} The Spatial Encoder is responsible for capturing the spatial dependencies between regions at any time interval $t-i$. It has four sub-components namely \textit{Local Feature Encoder (LFE)}, \textit{Semantic Feature Encoder (SFE)}, \textit{Temporal Feature Encoder (TFE)} and \textit{Global Feature Encoder (GFE)}.\\
\paragraph{Local Feature Encoder (LFE)} -- focuses on capturing local spatial dependencies. It utilises a \resnet{} \cite{resnet} to achieve this. Starting with the input image $\mathcal{X}_{t-i}$, it begins with an initial convolutional layer, increasing feature dimensions to $d$. This is followed by $m$ residual blocks designed to learn local neighbourhood relations. Each block consists of two convolutional layers and a skip connection. Finally, a $1 \times 1$ convolution merges feature maps, yielding the representation $\mathcal{Z}^{L}_{t-i}$ of the image:
\begin{equation}
    \mathcal{Z}^{L}_{t-i} = \resnet{}(\mathcal{X}_{t-i})  \in \mathbb{R}^{N\times d} \textnormal{,}
\end{equation}
where $N = H \times W$. Batch Normalisation \cite{batch_norm}, and GELU \cite{gelu} activation are applied after each convolution operation.

\paragraph{Semantic Feature Encoder (SFE)} -- Regions with similar semantic attributes exhibit spatial correlations similar to traffic features often occurring in proximity. For instance, recreational amenities like bars or pubs are often clustered in city centres for convenience. Therefore, another \resnet{} model, similar to \textit{LFE}, is employed to capture these semantic relationships and generate semantic representations: 
\begin{equation}
    \mathcal{Z}^{poi} = \resnet{}(\mathcal{X}^{poi}) \in \mathbb{R}^{N\times d} \textnormal{.}
\end{equation}

The residual blocks $m$ were set to $3$ for both \resnet{}s.

\paragraph{Temporal Feature Encoder (TFE)} -- generates the representation for temporal features utilising a multi-layer perception (MLP) containing two linear layers with GELU activation:
\begin{equation}
    \mathcal{Z}_{t-i}^{time} = \textsc{MLP}(\mathcal{X}_{t-i}^{time}) \in \mathbb{R}^{N\times d} \textnormal{.}
\end{equation}

\paragraph{Global Feature Encoder (GFE)} -- aims to capture the global dependency amongst distant regions utilising the Transformer's Encoder module. We first fuse all three features ($\mathcal{Z}^{L}_{t-i}$, $\mathcal{Z}^{poi}$ and $\mathcal{Z}_{t-i}^{time}$) along with our spatially constrained position embedding (described in Section \ref{sec:pos_embed}) that provides region position information through element-wise addition. The fused representation corresponding to each region $r^n$ is indicated by $z_{t-i}^{r^n}$.

Since not all regions hold equal significance at different intervals, our region sampling module (elaborated in Section \ref{sec:sampling}) then assigns importance weights to the fused representation of each region indicated by the colour darkness in the figure. Using these weights, only the top-$k$ regions are selected and utilised by the self-attention mechanism of the Transformer Encoder \cite{attention} to learn the global relation amongst the region and generate the final spatial representation for the image at time interval $t-i$:
\begin{equation}
    \mathcal{Z}_{t-i}^S = \transformer{}(z_{t-i}^{r^1}, \dots, z_{t-i}^{r^k}) \in \mathbb{R}^{d} \textnormal{.}
\end{equation}

\subsubsection{Temporal Encoder}

The Temporal Encoder captures the temporal dependency between images of different time intervals. Time is a crucial factor in determining the traffic trend; for instance, if it's late evening, traffic generally decreases as people commute back from work and vice versa in the early morning. Therefore, we first fuse the temporal features $ \mathcal{Z}_{t-i}^{time}$ with the spatial representation $\mathcal{Z}_{t-i}^S$ of images of input time intervals and then learn the temporal dependencies between them using another Transformer Encoder:
\begin{equation}
    \mathcal{Z}_{t+1}^{ST} = \transformer{}(\mathcal{Z}_{t-i}^S, \dots, \mathcal{Z}_{t}^S) \in \mathbb{R}^{d} \textnormal{,}
\end{equation}
where $\mathcal{Z}_{t+1}^{ST}$ is the spatio-temporal representation for the next time interval $t+1$.

\subsubsection{Predictor} The Predictor makes the final prediction utilising the spatio-temporal representation $\mathcal{Z}_{t+1}^{ST}$. It consists of a linear layer followed by $tanh(.)$ activation:
\begin{equation}
    \hat{\mathcal{X}}_{t+1} = tanh(\mathcal{Z}_{t+1}^{ST} \cdot \textbf{W} + \textbf{B})\textnormal{,}
\end{equation}
where $\textbf{W}$ and $\textbf{B}$ are the predictor's learnable parameters.

\subsection{Spatially Constrained Position Embedding}
\label{sec:pos_embed}
\begin{figure*}
    \centering
    \includegraphics[width=0.7\textwidth]{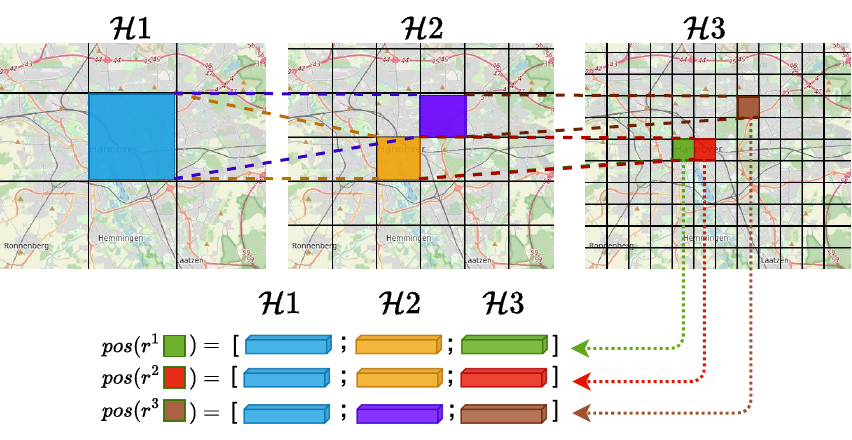}
    \caption{The city is divided into multiple levels of granularities ($\mathcal{H}_1$, $\mathcal{H}_2$, $\mathcal{H}_3$). Embeddings of different levels are concatenated to generate the position embedding (SCPE) of a region $r^n$. Map data: \copyright{} OpenStreetMap contributors, ODbL.}
    \label{fig:pos_embed}
\end{figure*}
%
We employ a hierarchical method for embedding the position information of regions, as illustrated in Figure \ref{fig:pos_embed}. To encode the position of a region $r^n$, we embed its position at multiple levels ($\mathcal{H}_1$, $\mathcal{H}_2$, $\mathcal{H}_3, \dots$). The lowest level captures the position at the highest granularity (e.g., $100 \textnormal{m} \times 100\textnormal{m}$), while the highest level captures the position at the lowest granularity (e.g., $5 \textnormal{km} \times 5\textnormal{km}$). The embeddings at different levels are concatenated to form the position embedding of the region $r^n$:
\begin{equation}
    pos(r^n) = [ \mathcal{H}_1 (r^n); \mathcal{H}_2 (r^n); \dots, \mathcal{H}_{\ell} (r^n)] \in \mathbb{R}^{d} \textnormal{,}
\end{equation}
where $\mathcal{H}_{\ell}(\cdot)$ represents the learnable embedding dictionary at level $\ell$. We set $\ell$ to $3$ in our model.

Semantically, SCPE captures hierarchical position information, including locality, city, state, country, etc.
Crucially, it informs the self-attention mechanism about each region's relative position, indicating proximity or distance. For example, in Figure \ref{fig:pos_embed}, regions $r^1$ and $r^2$ are immediate neighbours at two levels ($\mathcal{H}_1$ and $\mathcal{H}_2$), resulting in similar embedding vectors at those levels (depicted in blue and yellow). In contrast, $r^3$, a distant region, exhibits similarity at only one level $\mathcal{H}_1$ (depicted in blue). Consequently, this structure encourages the self-attention mechanism to prioritise attention to immediately adjacent neighbours while at the same time, its learnable design also allows it to learn the dependencies amongst distant neighbours.
Thus, our SCPE not only captures spatial dependencies at varying distances but also enhances semantic interpretation.

Our proposed position embedding has important application in diverse spatio-temporal tasks, specifically, when dealing with multiple geospatial points spanning various distances, from close (e.g., $10\textnormal{m}$) to very distant (e.g., $100\textnormal{km}$) such as trajectory representation learning \cite{traj_rep, traj_rep_2}, trajectory similarity comparison \cite{traj_sim, traj_sim_2} etc.

\subsection{Region Sampling}
\label{sec:sampling}
The importance of a region depends on its semantic attributes and the time of day. For instance, commercial regions are crowded during working hours and deserted during non-working hours. Therefore, we utilise semantic and temporal representations to predict an importance score for each region. Specifically, we first fuse $\mathcal{Z}^{poi}$ and $\mathcal{Z}_{t-i}^{time}$ and project them using an MLP consisting of a linear layer and GELU activation:
\begin{equation}
    \mathcal{Z}^{sampler} = \textsc{MLP}(\mathcal{Z}^{poi} + \mathcal{Z}_{t-i}^{time}) \in \mathbb{R}^{N \times d} \textnormal{,}
\end{equation}

To assign importance scores and prune regions, a naive approach involves mixing the representation of all regions using an MLP to generate a score for each region. However, this method increases the parameter count by $\mathcal{O}(N^2)$, making it inefficient.
Instead, we introduce a lightweight module that applies MLP individually to each region, thereby increasing the parameter count by $\mathcal{O}(N)$. To incorporate global information into the MLP, we split the sampler's representation ($\mathcal{Z}^{sampler}$) of each region $r^n$ into two parts: the first half is denoted as local features ($z^{r^n, local} \in \mathbb{R}^{d'}$), where $d' = \nicefrac{d}{2}$, and we compute the mean across all regions from the second half to obtain the global feature (${z}^{global} \in \mathbb{R}^{d'}$). We concatenate the local and global features and apply the MLP to compute the probabilities to keep or drop the region:

\begin{equation}
\begin{split}
{z'}^{r^n} = \textsc{MLP}([z^{r^n, local}, z^{global}]) \\
\rho = \textsc{Softmax}(\mathcal{Z'}) \in \mathbb{R}^{N \times 2} \textnormal{,}
\end{split}
\end{equation}
where $\rho^{r^n, 0}$ and $\rho^{r^n, 1}$ indicate the probabilities to drop and keep a region $r^n$, respectively.

Sampling directly from the distribution $\rho$ presents a challenge for end-to-end training due to its non-differentiability. To overcome this, we employ the Gumbel-Softmax technique \cite{gumbel_softmax}. This method is applied to the probability distribution of the regions to be kept, allowing us to sample important regions while ensuring the differentiability of the operation. Furthermore, Gumbel-Softmax introduces randomness to the sampling process, enabling the occasional selection of less important regions. This stochastic element enhances the model's robustness and its ability to generalize.

\subsection{Training}
To avoid training being dominated by large deviations, we utilise the mean squared error (MSE) together with the mean absolute error (MAE) as the loss function, which are defined as follows:
\begin{equation}
    \mathcal{L}_{MSE} = \frac{1}{N} \sum_{1}^{N}\left ( \mathcal{X}_{t+1} - \hat{\mathcal{X}}_{t+1} \right) ^ 2
\end{equation}
\begin{equation}
    \mathcal{L}_{MAE} = \frac{1}{N} \sum_{1}^{N} \left| \mathcal{X}_{t+1} - \hat{\mathcal{X}}_{t+1} \right|
\end{equation}

Furthermore, we use self-distillation to mitigate the information loss from region pruning at each time interval $t$. Our goal is to align the behaviour of the region-pruned student model with the teacher model containing all regions. It is achieved by minimizing the KL divergence \cite{kl_loss} between their spatial representations at each input time interval $t$:
\begin{equation}
    \mathcal{L}_{KL} = \frac{1}{T} \sum_{t=1}^T \textsc{KL}(\mathcal{Z}_{t}^{S, teacher} || \mathcal{Z}_{t}^{S, student}) \textnormal{,}
\end{equation}
where $T$ is the total number of input time intervals.

The overall training loss is the combination of the above three losses:
\begin{equation}
    \mathcal{L} = \mathcal{L}_{MSE} + \mathcal{L}_{MAE} + \alpha \cdot \mathcal{L}_{KL} \textnormal{,}
\end{equation}
where we set $\alpha = 0.3$ in all our experiments.

%% file: 05_setup.tex
\section{Evaluation Setup}
This section details our experiment setup, including the dataset, baseline models, and experimental settings.
\subsection{Datasets}
\input{tab_dataset}
We utilised floating car data from two German cities, Hannover and Dresden, as well as publicly available taxi data\footnote{https://www.nyc.gov/site/tlc/about/tlc-trip-record-data.page} from New York City (NYC). Each city was divided into $500 \textnormal{m} \times 500 \textnormal{m}$ regions. For the floating car data, we computed three types of measurements ($M=3$): \textit{Inflow} and \textit{Outflow} \cite{deepst}, representing the number of vehicles entering and exiting the region, respectively, and \textit{Density} \cite{deep_crowd}, measuring the number of vehicles within a region during a specific time interval. For NYC taxi data, we computed two types of measurements ($M=2$): \textit{Pickup} and \textit{DropOff} \cite{costnet}, indicating the number of taxis picked up and dropped off in a region, respectively. Dataset details are provided in Table \ref{tab:data}.

Additionally, we extracted $10$ categories of POIs from OpenStreetMap \cite{osm} to capture semantic characteristics. 
The POI categories and their counts for the three cities are described in Table~\ref{tab:poi}.
\input{tab_poi}

\subsection{Baselines}
\label{subsec:baselines}
We compare \approach{} to the following baselines:
\begin{itemize}
    \item \textbf{\textsc{Historical Average (HA)}} -- predicts by computing the average value of historical measurements.
    
    \item \textbf{\textsc{DeepST}} \cite{deepst} -- utilises CNNs to process the images corresponding to closeness, period, and trend, respectively. Subsequently, it employs element-wise addition to combine all the information. 

    \item \textbf{\textsc{ST-ResNet}} \cite{st-resnet} -- updates the CNN backbone of \textsc{DeepST} with that of the ResNet architecture.

    \item \textbf{\textsc{HConvLSTM}} \cite{hetro_convlstm} -- utilises ConvLSTM layers to jointly capture the spatio-temporal dependencies by only considering the closest historical time intervals to make the prediction.

    \item \textbf{\textsc{DeepSTN+}} \cite{deepstn+} -- is an upgrade of \textsc{ST-ResNet} using a \textit{ResPlus} unit that conducts two types of convolution: one to capture local neighbourhood relations with a smaller kernel size, and another to capture relations among remote regions with a larger kernel size.

    \item \textbf{\textsc{BDSTN}} \cite{bdstn} -- is a transformer-based architecture that first applies a multi-layer perception to learn spatial dependencies and then employs a transformer encoder to learn temporal dependencies.

    \item \textbf{\textsc{ST-GSP}} \cite{st_gsp} -- enhances \textsc{BDSTN} by replacing the MLP with that of a ResNet architecture for learning spatial dependencies.

    \item  \textbf{\textsc{DeepCrowd}} \cite{deep_crowd} -- stacks \textsc{ConvLSTM} layers in a pyramid architecture and fuses low and high-resolution feature maps to learn better feature representations.

    \item \textbf{\textsc{ST-3DGMR}} \cite{st3dgmr} -- utilises a dilated 3D Convolution \cite{conv3d} along with the residual connection to jointly capture the spatio-temporal relations.
    
    \item \textbf{\textsc{ST-3DMDDN}} \cite{st3dmddn} -- updates \textsc{ST-3DGMR} by breaking the 3D Convolution into 2D Convolution in spatial dimension and 1D Convolution in temporal dimension. This significantly lowers the number of parameters and increases the model's efficiency.
\end{itemize}

\subsection{Experimental Settings}
We designate December 2019 as the test set for Hannover and Dresden and December 2023 for New York. The remaining data of each city serves as the training set. To ensure model robustness, 20\% of the training data is used for validation. \approach{}'s feature dimension $d$ is set to 128. Following prior work \cite{deepstn+,st-resnet}, the historical time interval's closeness, period, and trend are set to $4$, $3$, and $2$, respectively to select historical spatio-temporal images. 

All models are trained on NVIDIA RTX A6000 GPUs using the AdamW optimiser \cite{adamw}. The learning rate is tuned within [$0.005$, $0.001$, $0.0005$, $0.0001$]. Training is conducted for $500$ epochs or until the validation error stagnates for $30$ consecutive epochs. Models achieving the best performance on the respective validation set are selected and evaluated on the test set. The effectiveness of the models is measured using root mean squared error (RMSE) and mean absolute error (MAE), whereas efficiency is measured using Giga Floating-Point Operations Per Second (GFLOPS)\footnote{We use the \texttt{calflops} \cite{calflops} module to compute GFLOPS of our model.}.

%% file: tab_dataset.tex
\begin{table*}[ht]
    \centering
    \caption{Dataset descriptions.}
    \begin{tabular}{lrrr}
    \toprule
          & \textbf{Hannover}  & \textbf{Dresden} & \textbf{NYC} \\
         \cmidrule{1-4}
         Latitude & $[52.3290, 52.4189]$ & $[51.0026, 51.0749]$ & $[40.7085, 40.8344]$ \\
         Longitude & $[9.6605, 9.8076]$ & $[13.6593, 13.8606]$ & $[-74.0214, -73.9146]$\\
         Grid size & $500 \textnormal{m} \times 500 \textnormal{m}$ & $500 \textnormal{m} \times 500 \textnormal{m}$ & $500 \textnormal{m} \times 500 \textnormal{m}$\\
         \#Regions & $400 = 20 \times 20$ & $448 = 16 \times 28$ & $504 = 18 \times 28$\\
         Time span & Jul - Dec 2019 & Jul - Dec 2019 & Jul - Dec 2023 \\
         Time Interval & 1 hour & 1 hour & 1 hour\\
    \bottomrule
    \end{tabular}
    \label{tab:data}
\end{table*}

%% file: tab_poi.tex
\begin{table}[ht]
    \centering
    \caption{Considered POI categories and their counts.}
    \begin{tabular}{lrrr}
    \toprule
          \multirow{2}{*}{\textbf{Category}}          & \multicolumn{3}{c}{\textbf{Count}} \\
        \cmidrule{2-4}
          & \textbf{Hannover}  & \textbf{Dresden} & \textbf{NYC}\\
         \cmidrule{1-4}
        Commercial & 1,008 & 1,618 & 752\\
        Culture & 557 & 418 & 1,100 \\
        Education & 306 & 339 & 655 \\
        Health & 513 & 698 & 891 \\
        Public Service & 486 & 572 & 823 \\
        Recreation & 1,314 & 1,256 & 7,362\\               
        Residential & 6,265 & 858 & 218 \\
        Sports & 882 & 923 & 1,740 \\
        Tourism & 869 & 1,178 & 1,687 \\
        Transport & 2,028 & 2,960 & 1,518\\
    \bottomrule
    \end{tabular}
    \label{tab:poi}
\end{table}

%% file: 06_result.tex
\section{Result}
This section presents \approach{}'s results, including baseline comparisons, sampling performance analysis, component contributions, spatial position embedding exploration, and a case study on sampling behaviour.

\subsection{Comparison with Baselines}
\input{tab_main_result}
The performance comparison of \approach{} and baselines on three datasets is detailed in Table~\ref{tab:main_result}.
The naive historical average (\textsc{HA}) performs the worst, emphasising the limitations of a simple heuristic-based approach in capturing the complex spatio-temporal dynamics of a city.

Among the deep learning baselines, \textsc{HConvLSTM} performs the poorest due to its limited consideration of only the closest historical information, disregarding periodicity and trend information. CNN-based approaches (\textsc{DeepST}, \textsc{ST-Resnet}, and \textsc{DeepSTN+}) perform better than \textsc{HConvLSTM} but exhibit inferior performance compared to \textsc{DeepCrowd}, \textsc{BDSTN}, and \textsc{ST-GSP}, attributed to their negligence of capturing temporal dependencies. 
The recently introduced 3D Convolution-based models \textsc{ST-3DGMR} and \textsc{ST-3DMDDN} exhibit the best performance among all baselines.

\approach{} outperforms the best baseline (\textsc{ST-3DMDDN}) by an average of $6.84\%$ w.r.t. RMSE and $6.75\%$ w.r.t. MAE across all the datasets, highlighting the importance of careful modelling all the three types of spatio-temporal relations in the architecture.

\subsection{Effect of Sampling}
\label{sec:result_sampling}
\begin{figure}[t]
    \centering
    \includegraphics[width=0.45\textwidth]{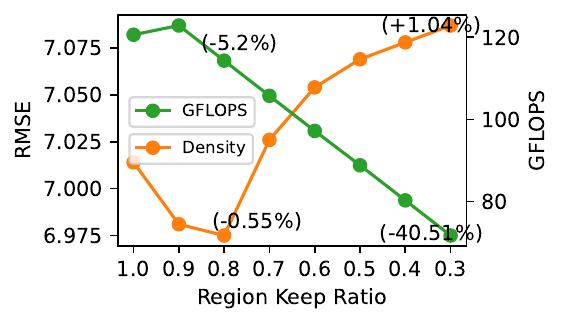}
    \caption{Effect of sampling on density prediction and corresponding computational cost for Hannover city. On x-axis is the ratio of region kept after sampling, on y-axis (left) is the RMSE for density prediction and on y-axis (right) is the corresponding GFLOPS.}
    \label{fig:gflops}
\end{figure}

Figure \ref{fig:gflops} depicts the effect of sampling on the performance of \approach{}. The values on the x-axis indicate the ratio of the regions kept after sampling. The values on the left and right axes indicate the RMSE in the density prediction of Hannover and GLOPS taken by the model. 

It can be seen that the computational cost monotonically decreases as the region keep ratio decreases except at the keep ratio of $0.9$. This is due to the addition of region sampling module parameters, which is not present on the base model when full attention is applied at a keep ratio of $1.0$.

Regarding performance, it can be seen that at first, the model gets better by approximately $0.55\%$ and then the performance degrades. This is because until a certain threshold ($0.8$ in this case), redundant and non-essential regions are pruned, thus, there is no loss of information. Additionally, distillation from the base model further regularises the model and makes it robust. Afterwards, the loss of information becomes significant, and performance degrades. To investigate this further, we also perform a case study which is described in Section \ref{sec:case_study}.

Overall, comparing effectiveness with efficiency, \approach{} can reduce the computational cost by approximately $\textbf{40\%}$ at the marginal drop in performance of around $\textbf{1\%}$.

\subsection{Contribution of Different Components}
\label{component_contrib}

In this study, we aim to investigate the contribution \approach{}'s different components for which we evaluated the following variants of \approach{} on the Hannover dataset:
\begin{itemize}

    \item \textbf{\textsc{w/o SCPE}} -- We replace our SCPE with a learnable position embedding similar to BERT \cite{bert} (details in Section~\ref{subsec:eval_scpe}).

    \item \textbf{\textsc{w/o \textit{LFE}}} -- In this variant, we replace \resnet{} of \textit{LFE} with that of a linear layer similar to BERT or ViT \cite{vit}.

    \item \textbf{\textsc{w/o \textit{SFE}}} -- We remove the \resnet{} of \textit{SFE}, and semantic features are not fed to the model.

    \item \textbf{\textsc{w/o \textit{TFE}}} -- Temporal features are not fed to the model in this variant.

    \item \textbf{\textsc{w/o \textit{GFE}}} -- We remove the transformer encoder of \textit{GFE} responsible for capturing global spatial dependency.

    \item \textbf{\textsc{w/o Temp. Enc.}} -- We remove the transformer encoder responsible for capturing temporal dependency, and the mean spatial representation of input intervals is fed to the predictor.
\end{itemize}
\input{tab_ablation}
The results in Table \ref{tab:ablation} demonstrate that each component contributes to the effectiveness of \approach{}. Specifically, the contributions of \textit{LFE}, \textit{GFE}, and \textit{Temporal Encoder} are the most important, emphasising the significance of capturing the local, global and temporal relation in spatio-temporal modelling.

\subsection{Learned Position Embedding}
\label{subsec:eval_scpe}

To further investigate the impact of Spatially Constrained Position Embedding (SCPE), we compare it with a learnable position embedding akin to \textsc{BERT} \cite{bert} as described in Section~\ref{component_contrib} (\textsc{w/o SCPE}). 
Results detailed in Table \ref{tab:ablation} demonstrate that passing neighbourhood information and constraining the model to pay more attention to closed neighbours \approach{}'s effectiveness. 

\begin{figure}[t]
\centering
\subfloat[All regions.\protect\footnotemark\label{fig:pos_embed_all}]{%
  \includegraphics[width=0.45\linewidth]{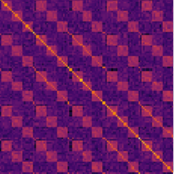}
}
\hfill
\subfloat[A specific region (bordered).\label{fig:pos_embed_region}]{%
  \includegraphics[width=0.45\linewidth]{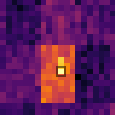}
}
   \caption{Visualisation of learned position embeddings in Hannover.} \label{fig:eval_pos_embed}
\end{figure}

\footnotetext{For better visualisation, we only show the correlation between $100$ regions in the north-western corner of the city.}

Furthermore, Figure \ref{fig:eval_pos_embed} depicts the correlation of the learned position embedding, where Figure \ref{fig:pos_embed_all} depicts how each region in the Hannover city correlates to another. 
The small square structures along the diagonal indicate that each region primarily attends to its immediate neighbours. Given that the regions are arranged row by row in this figure, the spaces before and after these squares represent the neighbouring regions of the rows above and below, respectively.
To better understand this, we also visualise the correlation of a specific region, illustrated in Figure \ref{fig:pos_embed_region}. 
This figure demonstrates how one region is related to the others in the city. It demonstrates the hierarchical nature where regions close to each other are more related than others. 

Thus, our SCPE, offering both semantic interpretability and improved predictive capability, proves beneficial for \approach{}.

\subsection{Case Study: Sampling Behaviour}
\label{sec:case_study}

\begin{figure*}[t]
\centering
\subfloat[Keep Ratio $1.0$.\label{fig:case_study1}]{%
  \includegraphics[width=0.3\linewidth]{images/case_study_gt.pdf}%
}%
\hfil
\subfloat[Keep Ratio $0.8$.\label{fig:case_study2}]{%
  \includegraphics[width=0.3\linewidth]{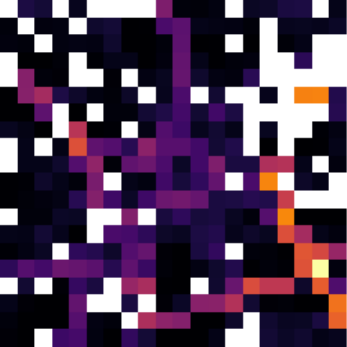}%
}
\hfil
\subfloat[Keep Ratio $0.5$.\label{fig:case_study3}]{%
  \includegraphics[width=0.3\linewidth]{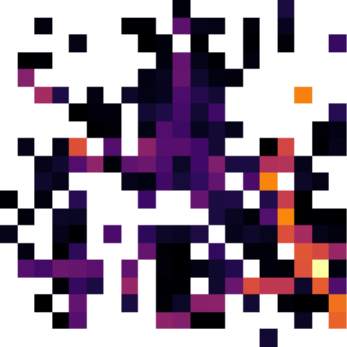}%
}%
   \caption{Case Study: Hannover traffic at 8-9 am on weekdays and the pruned regions (in white) at different keep ratios.} \label{fig:case_study}
\end{figure*}

To understand the sampling behaviour of our \approach{}, we performed a case study as illustrated in Figure \ref{fig:case_study}. Figure \ref{fig:case_study1} portrays the traffic patterns in Hannover between $8$ and $9$ am on a typical working day. Meanwhile, Figure \ref{fig:case_study2} and Figure \ref{fig:case_study3} exhibit the regions pruned to generate the representation of the same time interval at a keep ratio of $0.8$ and $0.5$, respectively.

Looking at the regions pruned at a ratio of $0.8$, it can be observed that \approach{} prunes two types of regions. The first category comprises regions with consistently negligible traffic, such as forested areas or regions containing lakes. The second category includes regions characterised by deterministic traffic patterns, exemplified by highways in the northeast of the city. Since these regions often contain redundant information, with traffic patterns mirroring those in other parts of the city, \approach{} considers them non-essential and prunes them. For instance, the traffic on pruned highways in the northeast can be accurately predicted based on the traffic of interconnected highways in the southeast and east. 

However, at the keep ratio of $0.5$, \approach{} tends to overprune, neglecting important non-redundant regions. Notably, regions towards the city centre, where traffic exhibits significant variability due to factors such as events, weather, etc., are overlooked. Consequently, the model's performance degrades, as discussed in Section \ref{sec:result_sampling} and illustrated in Figure \ref{fig:gflops} earlier.

%% file: tab_main_result.tex
\begin{table*}[ht]
\caption{\approach{}'s performance compared to the baselines in Hannover, Dresden and New York City. \textsc{Improv. (\%)} indicates the percentage improvement of \approach{} over the best-performing baseline (underlined).}
    \centering
    \addtolength{\tabcolsep}{-2.5pt} 
    
    \begin{tabular}{lrrrrrrrrrrrrrrrr}
         \toprule
         \multirow{3}{*}{\textbf{Approach}} & \multicolumn{6}{c}{\textbf{Hannover}} & \multicolumn{6}{c}{\textbf{Dresden}} & \multicolumn{4}{c}{\textbf{NYC}}\\
         \cmidrule{2-17}
         & \multicolumn{2}{c}{\textbf{Density}} & \multicolumn{2}{c}{\textbf{Inflow}} & \multicolumn{2}{c}{\textbf{Outflow}} & \multicolumn{2}{c}{\textbf{Density}} & \multicolumn{2}{c}{\textbf{Inflow}} & \multicolumn{2}{c}{\textbf{Outflow}} &
         \multicolumn{2}{c}{\textbf{Pickup}} & \multicolumn{2}{c}{\textbf{Dropoff}}
         \\
         \cmidrule{2-17}
         & \textbf{RMSE} & \textbf{MAE} & \textbf{RMSE} & \textbf{MAE} & \textbf{RMSE} & \textbf{MAE} & \textbf{RMSE} & \textbf{MAE} & \textbf{RMSE} & \textbf{MAE} & \textbf{RMSE} & \textbf{MAE} &
         \textbf{RMSE} & \textbf{MAE} & \textbf{RMSE} & \textbf{MAE}\\
         \midrule
         \textsc{HA} & 20.22 & 11.69 & 19.61 & 11.12 & 19.66 & 11.11 & 20.39 & 10.44 & 17.92 & 9.78 & 18.03 & 9.78 & 21.04 & 4.60 & 19.07 & 4.49 \\
         \textsc{HConvLSTM} & 9.27  & 5.54 & 9.28 & 5.50 & 9.19 & 5.43 & 9.30 & 5.29 & 9.00 & 5.16 & 8.88 & 5.10 & 8.36 & 1.67 & 6.92 & 1.49\\
         \textsc{DeepST} & 9.13 & 4.95 & 9.23 & 4.99 & 9.15 & 4.91 &  9.05 & 5.23  & 8.71 & 5.08 & 8.65 & 5.03 & 8.20 & 1.63 & 6.85 & 1.47\\
         \textsc{ST-ResNet} & 8.49 & 4.86 & 8.45 & 4.84 & 8.59 & 4.87 & 8.61  & 4.68 & 7.84 & 4.38 & 8.08 & 4.47 & 7.87 & 1.57 & 6.77 & 1.44\\
         \textsc{DeepSTN+} & 8.32 & 4.46 & 8.18 & 4.45 & 8.17 & 4.42 &  8.18 & 4.13  & 7.70 & 4.05  & 7.68 & 4.01 & 7.76 & 1.45 & 6.66 & 1.34\\
         \textsc{DeepCrowd} & 8.29 & 4.74 & 8.23 & 4.70 & 8.21 & 4.68 & 8.04 & 4.51 & 7.67 & 4.39 & 7.61 & 4.38 & 7.66 & 1.52 & 6.57 & 1.32\\
         \textsc{BDSTN} & 8.08 & 4.54 & 8.05 & 4.50 & 8.01 & 4.46 & 8.11  & 4.10 & 7.67 & 4.05 & 7.70 & 4.04 & 7.50 & 1.42 & 6.25 & 1.28\\
         \textsc{ST-GSP} & 7.90 & 4.46 & 7.93 & 4.43 & 7.90 & 4.41 & 7.93  & 4.01 & 7.66 & 4.01 & 7.69 & 4.00 & 7.24 & 1.40 & 6.07 & 1.29\\
         \textsc{ST-3DGMR} & 7.76 & 4.40 & 7.81 & 4.39 & 7.78 & 4.37 & 7.93 & 4.02 & 7.63 & 3.90 & 7.64 & 3.97 & 7.37 & 1.41 & 5.91 & 1.26\\
         \textsc{\underline{ST-3DMDDN}} & 7.62 & 4.31 & 7.72 & 4.32 & 7.67 & 4.29 & 7.75 & 3.91 & 7.59 & 3.88 & 7.60 & 3.94 & 7.03 & 1.37 & 5.84 & 1.25\\
         \approach{} & \textbf{7.01} & \textbf{4.00} & \textbf{7.11} & \textbf{4.03} & \textbf{7.10}  & \textbf{4.01} & \textbf{7.22} & \textbf{3.66} & \textbf{7.08} & \textbf{3.63} & \textbf{7.06} & \textbf{3.62} & \textbf{6.64} & \textbf{1.28} & \textbf{5.48} & \textbf{1.17}\\
         \midrule
         \textsc{Improv. (\%)} & 8.01 & 7.19 & 7.90 & 6.71 & 7.43 & 6.52 & 6.84 & 6.39 & 6.72 & 6.44 & 7.10 & 8.12 & 5.54 & 6.57 & 6.16 & 6.4 \\
         \bottomrule
    \end{tabular}
    \label{tab:main_result}
    \addtolength{\tabcolsep}{2.5pt} 
\end{table*}

%% file: tab_ablation.tex
\begin{table}[ht]
    \centering
    \addtolength{\tabcolsep}{-2pt}
    \caption{Contribution of different components of \approach{} on the Hannover dataset.}
    \begin{tabular}{lrrrrrr}
         \toprule
         & \multicolumn{2}{c}{\textbf{Volume}} & \multicolumn{2}{c}{\textbf{Inflow}} & \multicolumn{2}{c}{\textbf{Outflow}} 
         \\
         \cmidrule{2-7}
         & \textbf{RMSE} & \textbf{MAE} & \textbf{RMSE} & \textbf{MAE} & \textbf{RMSE} & \textbf{MAE} \\
         \midrule
         \approach{} & \textbf{7.01} & \textbf{4.00} & \textbf{7.11} & \textbf{4.03} & \textbf{7.10}  & \textbf{4.01} \\
         \textsc{w/o SCPE} & 7.17  & 4.07 & 7.27 & 4.09 & 7.25 & 4.07\\
         \textsc{w/o LFE} & 7.28  & 4.12 & 7.36 & 4.13 & 7.33 & 4.11 \\
         \textsc{w/o SFE}  & 7.16  & 4.01 & 7.29 & 4.06 &  7.24 & 4.03 \\
         \textsc{w/o TFE} &  7.12 & 4.05 & 7.20 & 4.07 & 7.18 & 4.05 \\
         \textsc{w/o GFE} & 7.22 & 4.10 & 7.34 & 4.13 & 7.32 & 4.11 \\
         \textsc{w/o Temp. Enc.} & 7.28 & 4.12 & 7.38 & 4.15 & 7.38 & 4.13\\
         \bottomrule
    \end{tabular}
    \label{tab:ablation}
    \addtolength{\tabcolsep}{2pt}
\end{table}

%% file: 07_relatedwork.tex
\section{Related Work}

Spatio-temporal prediction models have gained significant attention in the realm of smart city applications, leading to the development of numerous spatio-temporal models in recent years.

Early endeavours in this domain primarily focused on capturing local spatial relations. They utilised fully convolutional architectures where spatio-temporal images of input time intervals are stacked as channels and convolutions are used to capture both spatial and temporal dependencies, such as by \textsc{DeepST} \cite{deepst}, \textsc{ST-ResNet} \cite{st-resnet}, and \textsc{DeepSTN+} \cite{deepstn+}. 

However, these approaches tend to be inefficient in modelling complex temporal dependencies inherent in spatio-temporal data, relying solely on CNNs to capture them. Consequently, subsequent works proposed a two-component approach, separating the modelling of spatio-temporal dependency into spatial and temporal aspects. Works like \textsc{DMVST-Net} \cite{dmvst-net} and \textsc{STDN} \cite{stdn} pioneered this shift, utilising CNNs for spatial dependency and RNNs (LSTMs and GRUs) for temporal dependencies. 

The advent of \textsc{ConvLSTM} \cite{conv_lstm} and \textsc{ConvGRU} \cite{conv_gru} architectures further advanced spatio-temporal prediction techniques, evident in works like \textsc{Periodic-CRN} \cite{periodic_crn} and \textsc{DeepCrowd} \cite{deep_crowd}. Later, as Transformers \cite{attention} gained popularity for their superior ability to capture temporal dependency compared to LSTMs and GRUs, recent works utilised them to capture the temporal dependencies, as demonstrated by \textsc{BDSTN} \cite{bdstn} and \textsc{ST-GSP} \cite{st_gsp}.

More recent approaches draw inspiration from video representation learning \cite{video_representaion}, treating spatio-temporal data as sequences of frames and applying 3D Convolution to jointly learn spatio-temporal representations. Models such as \textsc{ST-3DGMR} \cite{st3dgmr} and \textsc{ST-3DMDDN} \cite{st3dmddn} exemplify this paradigm.

Unlike previous methods focusing on local neighbourhood relations via CNNs, our approach diverges by incorporating the self-attention mechanism to capture global relations. Furthermore, for efficiency, our sampling strategy mitigates the quadratic computational complexity associated with self-attention.

%% file: 08_conclusion.tex
\section{Conclusion}
\approach{} is a novel architecture for spatio-temporal modelling, effectively capturing essential spatio-temporal relations, including global spatial dependency often neglected in prior approaches.  
We propose a spatial position embedding which is semantically interpretable and enhances model performance.
Additionally, our region sampling strategy in \approach{} enhances efficiency by significantly reducing computation costs while maintaining performance.